\documentclass{article}

\PassOptionsToPackage{numbers}{natbib}
\usepackage[preprint]{neurips_2021}

\usepackage[utf8]{inputenc} 
\usepackage[T1]{fontenc}    
\usepackage{booktabs}       
\usepackage{amsfonts}       
\usepackage{nicefrac}       
\usepackage{microtype}      
\usepackage{amsmath}
\usepackage{graphicx}
\usepackage{tipa}
\usepackage{multirow}
\usepackage{longtable}

\usepackage{subcaption}
\usepackage{color}
\usepackage[table]{xcolor}
\usepackage{threeparttable}
\usepackage{diagbox}
\usepackage{enumitem}
\usepackage{tablefootnote}
\usepackage{hyperref}
\usepackage{url}            
\usepackage{xurl}
\usepackage{breakurl}

\usepackage[edges]{forest}
\usepackage{tikz}
\usepackage{tikz-qtree}
\usetikzlibrary{fadings}
\usetikzlibrary{mindmap,trees}
\usetikzlibrary{arrows,automata,shapes,positioning,shadows,trees}
\usetikzlibrary{shapes,snakes,shadows}
\usetikzlibrary{shapes.arrows}
\usetikzlibrary{calc,shapes, positioning}
\usetikzlibrary{decorations.text}
\usetikzlibrary{matrix,chains,positioning,decorations.pathreplacing,arrows}
\usetikzlibrary{bayesnet}
\usetikzlibrary{shadows.blur}
\usetikzlibrary{shapes.geometric}

\tikzstyle{edge}=[-latex',draw=black!90,shorten <=1pt,shorten >=1pt]
\tikzstyle{redge}=[latex'-,draw=black!90,shorten <=1pt,shorten >=1pt]
\tikzstyle{dedge}=[latex'-latex',draw=black!90,shorten <=1pt,shorten >=1pt]
\tikzstyle{block}=[draw, text width=5em,align=center,shape=rectangle, rounded corners, , align=center]
\tikzstyle{nobox}=[align=center]

\definecolor{emb}{RGB}{209,228,252}
\definecolor{hidden-blue}{RGB}{194,232,247}
\definecolor{hidden-orange}{RGB}{243,202,120}
\definecolor{hidden-yellow}{RGB}{242,244,193}
\definecolor{output-purple}{RGB}{219,203,231}
\definecolor{output-green}{RGB}{204,231,207}
\definecolor{hiddendraw}{RGB}{205, 44, 36}

\tikzstyle{mybox}=[
    rectangle,
    draw=hiddendraw,
    rounded corners,
    text opacity=1,
    minimum height=1.5em,
    minimum width=5em,
    inner sep=2pt,
    align=center,
    fill opacity=.2,
    ]

\tikzstyle{emb-purple}=[
    rectangle,
    draw=output-purple!50!purple,
    fill=output-purple,
    text opacity=1,
    minimum height=1.5em,
    minimum width=1.5em,
    inner sep=0pt,
    align=center,
    fill opacity=.9,
    ]

\tikzstyle{emb-blue}=[
    rectangle,
    draw=emb!50!blue,
    fill=emb,
    text opacity=1,
    minimum height=1.5em,
    minimum width=1.5em,
    inner sep=0pt,
    align=center,
    fill opacity=.9,
]

\definecolor{colone}{RGB}{178, 34, 34}
\definecolor{coltwo}{RGB}{106, 90, 205}
\definecolor{colthree}{RGB}{255, 250, 205}
\definecolor{colfour}{RGB}{0, 139, 69}
\definecolor{colfive}{RGB}{245,238,197}
\definecolor{colsix}{RGB}{243,235,179}
\definecolor{colseven}{RGB}{241,231,163}
\title{A Review of Deep Learning Based Image Super-resolution Techniques}

\author{
Fangyuan Zhu\\
\texttt{zhufangyuan@stu.xhu.edu.cn} \\
School of Computer and Software Engineering, Xihua University \\
}

\begin{document}
\maketitle

\begin{abstract}

Image super-resolution technology is the process of obtaining high-resolution images from one or more low-resolution images. With the development of deep learning, image super-resolution technology based on deep learning method is emerging. This paper reviews the research progress of the application of depth learning method in the field of image super-resolution, introduces this kind of super-resolution work from several aspects, and looks forward to the further application of depth learning method in the field of image super-resolution. By collecting and counting the relevant literature on the application of depth learning in the field of image super-resolution, we preliminarily summarizes the application results of depth learning method in the field of image super-resolution, and reports the latest progress of image super-resolution technology based on depth learning method.

\end{abstract}

\section{introduction}
\label{inrtoduction}
In recent years, artificial intelligence, as an important field of computer research, has achieved unprecedented great development. Among them, the application of deep learning method occupies an extremely important position. Deep learning methods are widely used in the field of image processing, such as image segmentation~\cite{shi2018key,li2013unsupervised,meng2013feature,yang2020new,meng2017weakly,meng2017seeds,meng2019weakly}, object detection~\cite{li2019headnet,qiu2020hierarchical,qiu2020offset}, image denoising, image dehazing~\cite{li2020region,wei2020single}, image deblurring and so on ~\cite{li2017research,liu2018image,cui2019pet,quan2021nonblind,albluwi2018image,zhao2020new,jiang2018medical,liao2019image}. In 2014, Dong et al.~\cite{dong2014learning} First applied the deep learning method to the field of image super-resolution, proposed a three-layer convolution neural network model for image super-resolution reconstruction, described it as a feature extraction layer, a nonlinear mapping layer and a reconstruction layer, and achieved the best results on the data set used at that time, It pioneered the application of depth learning method in the field of image super-resolution.

In this review, we mainly review the research work on five aspects, Upsampling in Image Super-resolution,Model Structure Design in Image Super-resolution, Cost Function in Image Super-resolution,Degradation Model in Image Super-resolution,and Common Datasets of Image Super-resolution. In section~\ref{Image Super-resolution}, and section~\ref{Deep Learning Technology} , we introduce the background and technical methods respectively. In section~\ref{Upsampling in Image Super-resolution}, We introduce several upsampling methods of image super-resolution. In section~\ref{Model Structure Design in Image Super-resolution}, We present several classical network architectures in image super-resolution. In section~\ref{Cost Function in Image Super-resolution}, we present several loss functions commonly used in image super-resolution. In section~\ref{Degradation Model in Image Super-resolution}, we show several common degradation models in image super-resolution. In section~\ref{Common Datasets of Image Super-resolution}, we present several of the most commonly used datasets in studies of image super-resolution respectively. Finally, in section~\ref{Conclusion}, we present conclusions based on the description and discussion in the above.

\tikzstyle{leaf}=[mybox,minimum height=1.2em,
fill=hidden-orange!50, text width=5em,  text=black,align=left,font=\footnotesize,
inner xsep=4pt,
inner ysep=1pt,
]

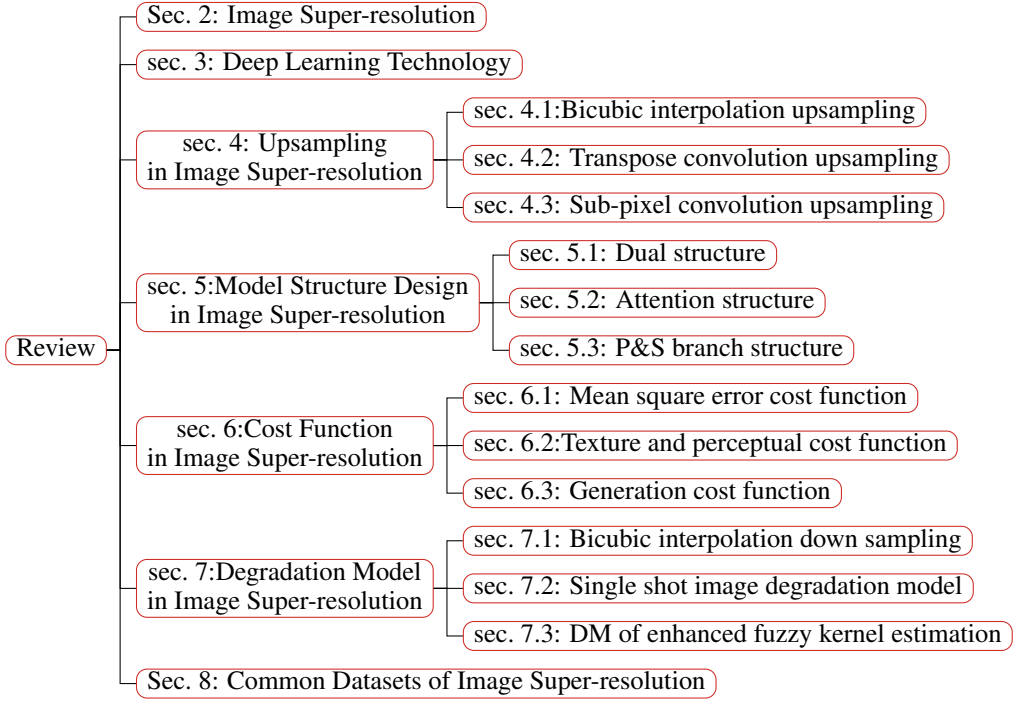
\begin{figure*}[thp]
 \centering
\begin{forest}
  forked edges,
  for tree={
  grow=east,
  reversed=true,  
  anchor=base west,
  parent anchor=east,
  child anchor=west,
  base=centering,
  font=\normalsize,
  rectangle,
  draw=hiddendraw,
  rounded corners,
  align=center,
  minimum width=2.5em,
  inner xsep=4pt,
  inner ysep=0pt,
  },
   [Review
                    [Sec.~\ref{Image Super-resolution}: Image Super-resolution
                    ]
                    [sec.~\ref{Deep Learning Technology}: Deep Learning Technology
                    ]
                    [sec.~\ref{Upsampling in Image Super-resolution}: Upsampling \\
                    in Image Super-resolution
                            [sec.~\ref{Bicubic interpolation upsampling}:Bicubic interpolation upsampling
                            ]
                            [sec.~\ref{Transpose convolution upsampling}: Transpose convolution upsampling
						    ]
                            [sec.~\ref{Sub-pixel convolution upsampling}: Sub-pixel convolution upsampling
                            ]
                    ]
                    [sec.~\ref{Model Structure Design in Image Super-resolution}:Model Structure Design \\
                    in Image Super-resolution
                            [sec.~\ref{Dual structure}: Dual structure
                            ]
                            [sec.~\ref{Attention structure}: Attention structure
                            ]
                            [sec.~\ref{Primary and secondary branch structure}: P\&S branch structure
                            ]
                    ]
       		    [sec.~\ref{Cost Function in Image Super-resolution}:Cost Function \\
                   in Image Super-resolution
                            [sec.~\ref{Mean square error cost function}: Mean square error cost function
                            ]
                            [sec.~\ref{Texture and perceptual cost function}:Texture and perceptual cost function
                            ]
                            [sec.~\ref{Generation cost function}: Generation cost function
                            ]
                    ]
			    [sec.~\ref{Degradation Model in Image Super-resolution}:Degradation Model\\ in Image Super-resolution
                            [sec.~\ref{Bicubic interpolation down sampling}: Bicubic interpolation down sampling
                            ]
                            [sec.~\ref{Single shot image degradation model}: Single shot image degradation model
                            ]
                            [sec.~\ref{Degenerate model of enhanced fuzzy kernel estimation}: DM of enhanced fuzzy kernel estimation
                            ]
                    ]
                    [Sec.~\ref{Common Datasets of Image Super-resolution}: Common Datasets of Image Super-resolution
                    ]
        ]
\end{forest}
\caption{Organization of this paper.}
\label{org_survey_paper}
\end{figure*}

\section{Image Super-resolution}
\label{Image Super-resolution}
\begin{figure}
    \centering
    \subfloat{
        \includegraphics[scale=0.5]{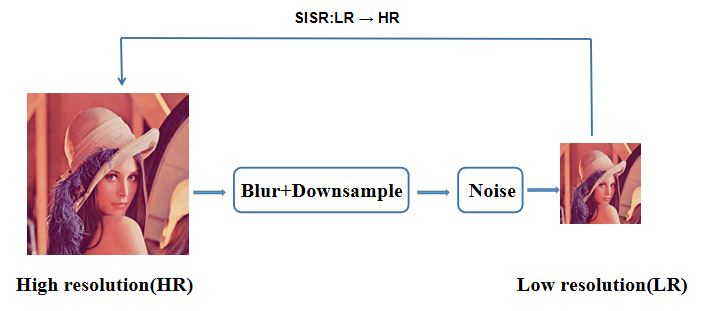}}
    \caption{schematic diagram of SISR.}
    \label{fig:introSISR}
\end{figure}

Image super-resolution refers to the image processing process of recovering a corresponding high-resolution image from a low-resolution image. According to the different number of input low resolution images, image super-resolution reconstruction can be divided into single image super-resolution reconstruction (SISR) and sequence image super-resolution reconstruction \cite{yue2016image}. To some extent, single image super-resolution reconstruction is the basis of sequence image super-resolution reconstruction, and the key of single image super-resolution reconstruction is to establish the relationship between low-resolution image and high-resolution image.
In the typical SISR framework, as depicted in Fig.~\ref{fig:introSISR}, the LR image y is modeled as follows:
\begin{equation}
    \begin{split}
        y = (x \otimes k){\downarrow}_{s} + n,
    \end{split}
    \label{defSISR}
\end{equation}
where $x \otimes k$ is the convolution between the blurry kernel $k$ and the unknown HR image $x$, ${\downarrow}_{s}$ is the downsampling operator with scale factor $s$, and $n$ is the independent noise term. Solving (\ref{defSISR}) is an extremely ill-posed problem because one LR input may correspond to many possible HR solutions. 

Image super-resolution was first proposed by Harris \cite{harris1964diffraction}and Goodman \cite{thompson1969modern} in the 1960s. Tsai \cite{tsai1989multiple} used multiple low-resolution images to restore high-resolution images in 1989. With the research and development of machine learning technology, Freeman et al.\cite{freeman2000learning} applied the method of machine learning to the field of image super-resolution for the first time in 2000. The evolution of image super-resolution technology has been summarized in several works \cite{park2003super,farsiu2004advances,yang2010image,gu2015convolutional,wang2015deep,zhang2019deep,glasner2009super,kim2010single,freeman2002example,capel2003computer,shi2014sub,elad2001fast,zhang2014super,nguyen2001computationally,molina2003parameter,akgun2005super,farsiu2004fast,vega2009super,yuan2010adaptive,patti1994high,elad1997restoration,suo2011robust,sun2008image}.

Generally speaking, through the transformation of the hardware part of the imaging system, the performance of the imaging system and image resolution can be improved to a certain extent. There are usually three ways to transform the hardware part of the imaging system: (1) reduce the size of the pixel sensor, that is, increase the number of pixels on the sensor per unit area in the imaging device. However, as the size of the sensor decreases, the effective light intensity per unit pixel decreases, resulting in image noise. (2) Increasing the chip size can increase the number of pixels, but the increase of chip size will increase the capacitance and affect the charge transfer rate. (3) By increasing the focal length of the camera to enhance the spatial resolution of the image, however, this method will bring negative effects such as the increase of the volume and weight of the imaging equipment and the size of the optical components, which greatly improves the manufacturing difficulty and cost of optical materials. Due to the above reasons, image super-resolution technology is mostly studied from the software technology level of image processing methods.

The process of image super-resolution reconstruction is still a serious mathematical underdetermination problem. It is mathematically impossible to obtain a unique high-resolution image from a low-resolution image without any prior constraints. Because the image degradation process from high-resolution image to low-resolution image shows that there are often multiple high-resolution images, and the same low-resolution image can be obtained through degradation. This makes the image super-resolution reconstruction process become a serious underdetermined process. Because of this characteristic of image super-resolution reconstruction process, in recent years, researchers in this field mainly use learning based methods to learn image prior information from a large number of data to restrict the solution space, so as to obtain the optimal solution of the problem.

The traditional image super-resolution reconstruction algorithms mainly include the following three categories: interpolation based image super-resolution, reconstruction based image super-resolution and learning based image super-resolution. The image super-resolution technology based on interpolation includes nearest neighbor interpolation, bilinear interpolation, cubic interpolation and so on. The main idea is to calculate the value of this point according to a certain formula through the values of several known points around a point and the positional relationship between the surrounding points and this point, so as to improve the resolution. Generally speaking, the interpolation algorithm improves the image detail is limited, so it is less used. Generally speaking, reconstruction by interpolation algorithm between multiple images is a means. In addition, in video super-resolution reconstruction, by interpolating and adding new frames between two adjacent frames, the video frame rate can be improved and the sense of picture frustration can be reduced. Image super-resolution technology based on reconstruction is usually based on multi frame images, which needs to combine a priori knowledge. Generally, starting from the image degradation model, it is assumed that the low-resolution image is obtained from the high-resolution image through motion transformation, blur and noise. These methods mainly include convex set projection method, Bayesian analysis method, iterative back projection method, maximum a posteriori probability method, regularization method, hybrid method and so on. Learning based image super-resolution technology mainly uses a large number of pre training data to learn the mapping relationship between low-resolution image and high-resolution image, and then predict the high-resolution image corresponding to the low-resolution image according to the learned mapping relationship, so as to realize the super-resolution reconstruction process of the image. Common learning based methods include neighborhood embedding method, support vector regression method, manifold learning, sparse representation and so on. Sparse representation is mainly based on compressed sensing theory. Compressed sensing theory means that an image can be accurately reconstructed from a set of sparse representation coefficients in an ultra complete dictionary under very harsh conditions. Through the joint training of low-resolution image block dictionary and high-resolution image block dictionary, we can strengthen the similarity between low-resolution and high-resolution image blocks and their corresponding real dictionary sparse representation, so that the sparse representation of low-resolution image blocks and high-resolution super complete dictionary can reconstruct high-resolution image blocks, Then, the final complete high-resolution image is obtained by connecting the high-resolution image blocks. Learning dictionary pair is a more compact representation of image block pairs. It only needs to sample a large number of image block pairs. Compared with traditional methods, the computational cost of this method is significantly reduced. The effectiveness of sparse representation is proved in the special cases of image super-resolution reconstruction and face illusion. In these two cases, the high-resolution image generated by sparse representation is highly competitive, and even has more advantages than other similar image super-resolution methods in the quality of the generated image. In addition, the local sparse model of sparse representation method has adaptive robustness to noise. Therefore, the sparse representation method can perform image super-resolution processing on noisy input images in a unified framework.

\section{Deep Learning Technology}
\label{Deep Learning Technology}

Deep learning is a broader class of machine learning methods based on data representation. It combines low-level features to form more abstract high-level representation features to find the distributed features of data. Deep learning can represent more and more abstract concepts or patterns level by level. Take an image as an example, its input is a pile of original pixel values. In the depth learning model, the image can be expressed level by level as the edge of a specific position and degree, the pattern obtained by the combination of edges, the pattern of a specific part obtained by the further combination of multiple patterns, and so on. Finally, the model can easily complete a given task according to a more level representation, such as identifying objects in the image. It is worth mentioning that as a kind of representation learning, deep learning will automatically find out the appropriate way to represent data at each level.

Deep learning appears as a branch of machine learning. The main goal of machine learning is to improve the performance of the system by allowing computers to learn from historical experience. There are obvious differences between deep learning and traditional machine learning methods. Deep learning technology emphasizes the use of multi-layer neural network cascade for feature extraction and representation.

It is generally believed that the rise of the technology wave of deep learning began around 2010. It is the first successful attempt to apply deep convolution network to the field of image recognition in alexnet network. Around 2011, researchers applied deep learning technology in the field of speech recognition and made a major breakthrough. In 2016, alphago developed by deepmind based on deep learning technology defeated the world go champion, making the deep learning technology widely available to the public.

Deep learning has a very obvious external characteristic, that is, it can realize end-to-end training. Different from the traditional machine learning model, the implementation of deep learning model is usually not to piece together individual functional modules to form a system, but to optimize the parameters by global training after the whole system is designed. In the field of image vision, most of the previous processing models tend to process feature extraction separately from the construction of machine learning model. After the application of deep learning technology, the feature extraction process will usually become a part of the whole network model and be replaced by the automatically optimized step-by-step convolution kernel. In addition to this, deep learning technology has some other characteristics different from the previous classical machine learning methods, including the inclusion of non optimal solutions and the use of non convex nonlinear optimization.

\section{Upsampling in Image Super-resolution}
\label{Upsampling in Image Super-resolution}

From a certain point of view, the image super-resolution problem can be decomposed into two subproblems. One of them is the enlargement of image size, which is commonly referred to as image sampling. The up sampling methods used in image super-resolution technology based on depth learning usually include the following categories:

\subsection{Bicubic interpolation upsampling}
\label{Bicubic interpolation upsampling}

Bicubic interpolation upsampling is a relatively traditional upsampling method. The original depth super division method srcnn embeds bicubic interpolation upsampling in the front end of the network. At the beginning of the network, it first enlarges the size of the low resolution image, and then uses the convolution layer to extract the image features, The size of the final output high-resolution image is consistent with that after bicubic interpolation amplification at the beginning. This low resolution image upsampling method in the hyperspectral network is later also called pre upsampling, which makes the acquisition of high-resolution images with different magnification completely depend on the selection of bicubic interpolation magnification at the beginning of the hyperspectral network. This upsampling strategy is also used by the super sub network in \cite{kim2016deeply,tai2017image,sajjadi2017enhancenet,tai2017memnet}.

\subsection{Transpose convolution upsampling}
\label{Transpose convolution upsampling}
Transpose convolution up sampling is an image up sampling method based on convolution operation. Transpose convolution up sampling is used in FSRCNN \cite{dong2016accelerating} to obtain the final high-resolution output picture. Because transpose convolution enlarges the size of low resolution pictures through convolution operation, considering that this operation will greatly increase the amount of calculation of the network in the super division network, transpose convolution up sampling is generally embedded at the end of the super division network, forming an up sampling method later called post up sampling. In addition to adopting such an up sampling structure in FSRCNN, the super sub network in \cite{tong2017image,li2019feedback} also uses such an image up sampling method.

\subsection{Sub-pixel convolution upsampling}
\label{Sub-pixel convolution upsampling}
Sub-pixel convolution up sampling is also an image up sampling method based on convolution operation. Different from transpose convolution, sub-pixel convolution up sampling generates multiple image feature channels through convolution, and up sampling of images is realized through feature channel shaping. This upsampling method can often achieve better results in super segmentation tasks because of its high utilization of the information of low-score pictures. Like transpose convolution, sub-pixel convolution will also increase the amount of calculation of the network. In order to achieve faster super division speed, sub-pixel convolution up sampling is usually embedded at the end of the super division network to achieve a better balance between super division performance and operation efficiency. The sub-pixel convolution up sampling image up sampling method is adopted in the super division networks in \cite{ledig2017photo,lim2017enhanced,zhang2018residual,ahn2018fast,li2018multi,zhang2018image,wang2018esrgan,zhang2019residual,dai2019second}.

\tikzstyle{leaf}=[mybox,minimum height=1.2em,
fill=hidden-orange!50, text width=5em,  text=black,align=left,font=\footnotesize,
inner xsep=4pt,
inner ysep=1pt,
]

\begin{figure*}[thp]
    \centering

    \begin{forest}
        forked edges,
        for tree={
                grow=east,
                reversed=true,  
                anchor=base west,
                parent anchor=east,
                child anchor=west,
                base=left,
                font=\normalsize,
                rectangle,
                draw=hiddendraw,
                rounded corners,
                align=left,
                minimum width=2.5em,
                inner xsep=4pt,
                inner ysep=0pt,
            },
        where level=1{text width=6em,font=\normalsize}{},
        where level=2{text width=6.9em,font=\normalsize}{},
        where level=3{font=\footnotesize,yshift=0.25pt}{},
        [Upsampling in Image Super-resolution
       	[Bicubic \\
        upsampling
        	[DRCN~\cite{kim2016deeply}{,}DRRN~\cite{tai2017image}\\
        EnhanceNet~\cite{sajjadi2017enhancenet}{,}MemNet~\cite{tai2017memnet}
        ,leaf,text width=12em
        	]
        ]
        [Transpose \\
        convolution\\
        upsampling
           [FSRCNN~\cite{dong2016accelerating}{,}SRDenseNet~\cite{tong2017image}\\
                SRFBN~\cite{li2019feedback}
                ,leaf,text width=12em
            ]
        ]
  		[Sub-pixel \\
        convolution \\
        upsampling
            [SRGAN~\cite{ledig2017photo}{,}EDSR~\cite{lim2017enhanced}\\
             RDN~\cite{zhang2018residual}{,}CARN~\cite{ahn2018fast}\\
                MSRN~\cite{li2018multi}{,}RCAN~\cite{zhang2018image}
                ,leaf,text width=12em
            ]
        ]
        ]
    \end{forest}
    \caption{A classification of Upsampling in Image Super-resolution.}
    \label{main_taxonomy_of_Upsampling in Image Super-resolution}
\end{figure*}
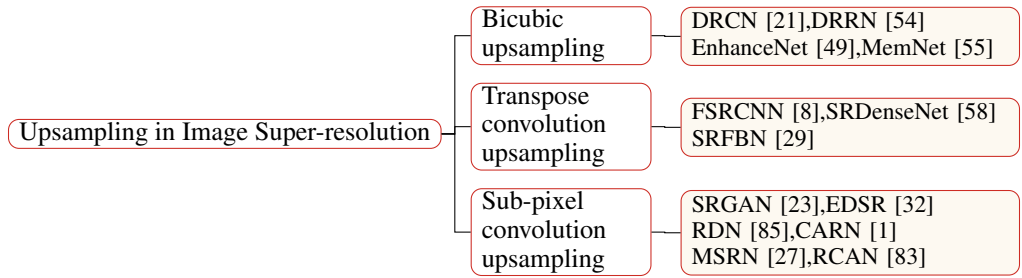

\section{Model Structure Design in Image Super-resolution}
\label{Model Structure Design in Image Super-resolution}
One of the most extensive variants of image super-resolution technology based on deep learning method is the design of model structure. At present, the common super sub network model structures generally include the following categories:

\subsection{Dual structure}
\label{Dual structure}
Dual structure regards super-resolution as a closed-loop problem. Dual super-resolution network generally includes two branches: low score to high score process and high score to low score process. DRN \cite{guo2020closed} is a typical dual structure super-resolution network. In DRN, the process from low score to high score is a u-net network that down samples first and then up samples. In the down sampling process, the network can extract the local feature information in the picture to the greatest extent. In the up sampling process, the network fuses the local feature information with the global feature information. This structure ensures that the process of super-resolution reconstruction can make full use of the shallow and deep information of the image. The disadvantage is that the information of different scale levels has not been effectively fused. The process from high score to low score is a down sampling process, which can learn that the reconstruction result degenerates into a function of low score image, which imposes strong solution space constraints on the reconstruction process. Through the joint training of the two branches, the size of the solution space can be reduced and the super-resolution performance can be enhanced.

\subsection{Attention structure}
\label{Attention structure}
Attention structure regards super-resolution as a sub regional image restoration problem. Attention structure super-resolution usually designs the corresponding attention module for each part of the image. CDC \cite{wei2020component} proposed a divide and conquer attention structure super-resolution model. CDC network is essentially a specially designed attention network, but the object of attention in the network is neither channel nor feature image pixel space, but flat area, edge area and corner area separated from high score image by using Harris corner detection algorithm. The network consists of three component attention modules. Each component attention module generates an attention mask and intermediate SR result. The final reconstruction result is generated by multiplying and summing the attention mask and intermediate SR result of each module. CDC applies the attention mechanism to different contents in the image, so that the network gives greater weight to the pixels that are more beneficial to the super segmentation reconstruction, and improves the super segmentation performance. However, the corner detection algorithm used by CDC will lead to the decline of the running speed of the algorithm. It is still a problem worth exploring whether we can learn the image regions with different importance for super-resolution reconstruction through the network.

\subsection{Primary and secondary branch structure}
\label{Primary and secondary branch structure}
The primary and secondary branch structure emphasizes the use of secondary branch tasks to help improve the performance of primary branch over sub tasks. SPSR \cite{ma2020structure} is a super sub network with primary and secondary branch structure. SPSR believes that the image structure information is very important information for super-resolution reconstruction. However, the previous super-resolution network mining the image structure information is still very limited. Therefore, the author proposes a structure preserving super-resolution network. Specifically, the SPSR super-resolution network has two branches. The first branch uses the existing traditional super-resolution network to complete the task of super-resolution reconstruction, and the second branch is a separate branch of image gradient information mining. The image gradient information extracted by the gradient information extraction branch is used as the main branch of the image a priori to assist the super division reconstruction, so as to improve the performance of the super division reconstruction. Experiments show that the gradient information learning module proposed by the author effectively enhances the processing ability of the network to image structure and texture, and improves the reconstruction performance. However, whether we can mine other image prior information to assist the image super segmentation reconstruction process and improve the super segmentation performance is still a problem to be explored.

\tikzstyle{leaf}=[mybox,minimum height=1.2em,
fill=hidden-orange!50, text width=5em,  text=black,align=left,font=\footnotesize,
inner xsep=4pt,
inner ysep=1pt,
]

\begin{figure*}[thp]
    \centering

    \begin{forest}
        forked edges,
        for tree={
                grow=east,
                reversed=true,  
                anchor=base west,
                parent anchor=east,
                child anchor=west,
                base=left,
                font=\normalsize,
                rectangle,
                draw=hiddendraw,
                rounded corners,
                align=left,
                minimum width=2.5em,
                inner xsep=4pt,
                inner ysep=0pt,
            },
        where level=1{text width=6em,font=\normalsize}{},
        where level=2{text width=6.9em,font=\normalsize}{},
        where level=3{font=\footnotesize,yshift=0.25pt}{},
        [Model Structure Design \\
        in Image Super-resolution
        	[Dual structure
       	]
        	[Attention \\
            structure
        	]
  		    [Primary  and \\
            secondary \\
            branch structure
        	]
        ]
    \end{forest}
    \caption{A classification of Model Structure Design in Image Super-resolution.}
    \label{main_taxonomy_of_Model Structure Design in Image Super-resolution}
\end{figure*}
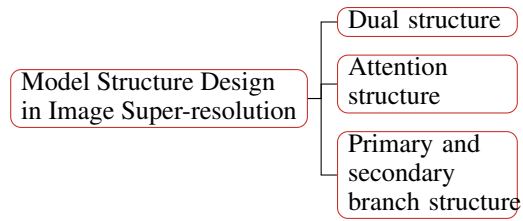

\section{Cost Function in Image Super-resolution}
\label{Cost Function in Image Super-resolution}
The design of cost function also has an important impact on the effect of image super-resolution. The cost function is an important indicator to drive the super-resolution model to obtain the desired high-resolution picture. The cost functions used in the image super-resolution model based on depth learning method roughly include the following categories:

\subsection{Mean square error cost function}
\label{Mean square error cost function}
Mean square error (MSE) is a relatively traditional but widely used cost function. The mean square error is a cost function per pixel, and the mean square error is used as the cost function in RTSR \cite{park2021recurrently}. RTSR believes that the quality of the high score image as the supervision image can be continuously improved through the trained super score network, so as to continuously supervise the training of the original super score network with the high score image with better quality, so as to improve the performance of the super score model. RTSR designs a two-stage hyper divided network model, and obtains the final network training results through multiple cycles of the two stages. Specifically, the first stage is the traditional super fractional reconstruction process, which takes the low-grade pictures as the input, and completes the network training by reducing the loss between the reconstruction results and the supervised pictures. The second stage is to monitor the image generation process with better quality. By inputting the original high score image into the super sub network trained in the first stage, we can get the enhanced supervision image with better quality. It is noted that the process of generating the reconstruction result by using the super sub network is accompanied by the enlargement of the picture size, so a down sampling module is specially embedded in the network in the second stage to reduce the size of the generated supervision picture to the same size as the original high score image. RTSR improves the performance of the super sub network by enhancing the quality of supervision pictures and repeatedly training the network. However, whether using enhanced supervision pictures to continuously improve the quality of network training will cause more artificial traces in the final generated pictures is worth further research, In addition, it is also worth exploring whether the supervision picture enhanced in the last cycle can be used to directly generate the supervision picture required in the next cycle instead of continuously using the original high score picture to generate the supervision picture required in the next cycle.

\subsection{Texture and perceptual cost function}
\label{Texture and perceptual cost function}
Texture loss is used to describe the difference in texture style between the generated image and the reference image. Perceptual loss can evaluate the difference of perceptual quality between different images. Specifically, fully trained natural image classification models, such as VGg and RESNET, are used to extract the features of different images, and then calculate the perceptual distance.

\subsection{Generation cost function}
\label{Generation cost function}
In the super-resolution network model based on GAN, it is usually combined with anti loss on the basis of pixel by pixel loss or perceptual loss. The purpose is to enable the discriminator to extract potential patterns that are difficult to learn from the real reference image through the competition between the generator and the discriminator, and force the generator to adjust the model, This enables the generator to produce realistic high-resolution images. Because the training of GAN is still difficult and unstable at this stage, the super-resolution model combined with anti loss sometimes produces artificial traces and unnatural deformation. How to better apply GAN to the field of image super-resolution reconstruction is still worthy of in-depth research.

\section{Degradation Model in Image Super-resolution}
\label{Degradation Model in Image Super-resolution}

Image degradation is the opposite process of image super-resolution. The process of image degradation is the process of image information loss, and the process of image super-resolution is the process of image information expansion. Because the image degradation process is irreversible, the image super-resolution process can only approximate the opposite process as much as possible to obtain the approximate value of the high-resolution image before degradation. Selecting the image degradation model closer to the real image acquisition scene can make the approximate value obtained by the image super-resolution model closer to the high-resolution image in the real scene. The degradation models used in the image super-resolution model based on depth learning method generally include the following categories:

\subsection{Bicubic interpolation down sampling}
\label{Bicubic interpolation down sampling}
Bicubic interpolation down sampling is a relatively traditional image degradation method. The original depth super division method srcnn uses bicubic interpolation down sampling to degrade high-resolution images to obtain low-resolution images required for super-resolution network input. Then, at the beginning of the network, the size of the low-resolution image is enlarged, then the convolution layer is used to extract the image features, and finally the high-resolution image is output. This image degradation method is also used by many subsequent image super-resolution models. However, this makes the network constantly approximate the inverse process of bicubic down sampling process, and its ability to adapt to the inverse process of image degradation process in real scene is limited. If we continue to use double triple up sampling to enlarge the image size in the super-resolution network model, the loss of image information caused by the two operations will reduce the performance of the super-resolution network.

\subsection{Single shot image degradation model}
\label{Single shot image degradation model}
Due to the irreversibility of the real image degradation process, the image super-resolution problem is a serious ill posed ill posed problem. The relationship between high-resolution image and low-resolution image is non injective. IRN \cite{xiao2020invertible} proposed to model the down sampling process of the image as a reversible injective process. It reduces the morbid nature of the sampling process on the image. In the down sampling process, the high score image is processed into a low score image and sample independent auxiliary variables containing high-frequency information by wavelet transform and reversible neural network. In the up sampling process, the auxiliary variable Z and low score image are processed by reversible neural network and inverse wavelet transform to obtain the reconstruction results. The whole network training process forms a bijective function. The network enhances the processing of the lost information in the down sampling process and improves the reconstruction performance of SR network.

\subsection{Degenerate model of enhanced fuzzy kernel estimation}
\label{Degenerate model of enhanced fuzzy kernel estimation}
The classic image super-resolution degradation model describes the image degradation process as a process of fuzzy kernel convolution, image down sampling and superimposed noise. Relevant studies have also confirmed the importance of fuzzy kernel for image super-resolution results. FKPSR \cite{liang2021flow} is a further deepening of the SR reconstruction work of blind super-resolution. FKPSR mainly proposes a network architecture using standardized flow for fuzzy kernel estimation. Considering the importance of fuzzy kernel to the performance of super-resolution reconstruction, FKPSR enhances the representation and generalization ability of fuzzy kernel carried by blind super-resolution network for various low-resolution images by realizing the accurate estimation of fuzzy kernel, and improves the performance of blind super-resolution reconstruction. This is of great significance for the practical application of super division reconstruction in real images. However, the noise and down sampling factors in the classical degradation model also have a certain restrictive effect on the performance improvement of blind super division reconstruction. How to better model these priors to improve the performance of blind super division network remains to be explored.

\section{Common Datasets of Image Super-resolution}
\label{Common Datasets of Image Super-resolution}
Supervised image super-resolution data sets are mainly divided into two categories. One of them is the data set synthesized manually. The synthesis methods mainly include bicubic interpolation and so on. Set5, set14, B100, urban100, manga109, div2k, etc. all belong to this kind of data set. Another kind of data set is the data set obtained by shooting with a variety of different camera devices under different scale factors, or the data set synthesized by using a variety of fuzzy cores and noise. This kind of data set is considered to be closer to the image degradation process in the real scene, Thus, the trained super-resolution model can be more suitable for image super-resolution tasks in real scenes. Whether it is the data set synthesized by interpolation or the data set generated by fuzzy kernel degradation, the quality evaluation of the generated high-resolution image is very important for the image super-resolution task. \cite{wu2015blind,wu2015highly,ma2016group,wu2018perceptually,wu2017blind1,wu2017blind2,wu2015no,wu2020subjective,tang2017deep,wu2014no,meng2019new,wu2017generic,wu2019beyond,wu2016gip} discusses several methods for image quality evaluation, which is of great significance to explore the cost function with better performance in the field of image super-resolution.

\section{Application of deep learning in image super-resolution}
\label{Application of deep learning in image super-resolution}
The application of deep learning technology has touched many research fields. Deep learning has been widely used in the field of image vision. Since Dong et al. First applied convolutional neural network to the research of image super-resolution in 2014, a large number of new image super-resolution methods based on depth learning have been emerging. These methods can be roughly divided into the following categories: image super-resolution based on convolutional neural network, image super-resolution based on residual network and dense network, and image super-resolution based on GAN network. These three methods are introduced below.

The main network structures of image super-resolution algorithm based on convolutional neural network are srcnn and FSRCNN. The performance of this method is much better than the previous algorithms. However, because the network layer is too few and the receptive field of the network is relatively small, the reconstructed image still has many defects in detail and texture, and the reconstruction effect of the high-frequency part of the image is poor. And srcnn is not a real end-to-end processing method. It needs bicubic interpolation to enlarge the low resolution image to the size of the target image before the low resolution image is input into the network, and then enter the network model for high-resolution image reconstruction. This process will make srcnn need to process the image in high-resolution space, resulting in an increase in the amount of calculation. FSRCNN algorithm realizes end-to-end reconstruction from low resolution image to high resolution image by using deconvolution layer.

Image super-resolution algorithm based on residual network and dense network. The main network models are RESNET network \cite{kim2016accurate} and densenet network. RESNET network adopts a more conservative way to deepen the depth of the network, and has achieved good results in the deep network. The general understanding of residual structure holds that each residual block learns new information on the basis of maintaining the original characteristics as much as possible. RESNET structure is similar to densenet structure. It can reuse the convolution layer features at all levels, and the parameters of the reuse layer will not be increased in the process of feature reuse. Densenet structure can explore new features more effectively than RESNET in the process of feature reuse, but densenet's cascade form will make the network parameters surge with the increase of layers, which is not conducive to the construction of practical network structure. Densenet inputs the features of each layer to all subsequent layers in a dense block, so that the features of all layers are connected in series, rather than adding directly like RESNET. This structure brings the advantages of reducing the gradient disappearance problem, strengthening feature propagation, supporting feature reuse and reducing the number of parameters to the whole network. Densenet applies the dense block structure to the super-resolution problem. Its structure can be divided into four parts: (1) learning the characteristics of the lower layer with a convolution layer; (2) Learning high-level features with multiple dense blocks; (3) Up sampling through several deconvolution layers; (4) High resolution output is generated through a convolution layer. Dense block divides many convolution networks into 3 ~ 4 parts, which are called dense block. In each block, dense links are executed, and the output of the block is pooled to reduce the size of the feature graph.

The representative network structures of image super-resolution algorithms based on GAN network are SRGAN and ESRGAN. The core idea of SRGAN network comes from the Nash equilibrium of game theory. The network assumes that the two parties participating in the game are respectively a generative model and a discriminant model. The goal of the generative model is to learn as much as possible the real data distribution, and the goal of the discriminant model is to try to correctly judge whether the input data is real data or from the generative model. In order to win the game, two game participants need to continuously optimize their respective parameter settings to improve their own generation ability and discrimination ability. This learning optimization process is to find a Nash equilibrium between the two. However, the enlarged details of the SRGAN network are usually accompanied by artifacts. In order to further improve the visual quality, the author carefully studied the three key parts of SRGAN:network structure;adversarial loss;perceptual domain loss. And improve each item to get ESRGAN. ESRGAN proposes a Residual-in-Residual Dense Block network unit, in which the BN layer is removed. In addition, ESRGAN draws on the idea of Relativistic GAN, allowing the discriminator to predict the authenticity of the image rather than whether the image is a fake image. ESRGAN also improves the perceptual domain loss, using features before activation, which can provide stronger supervision for brightness consistency and texture restoration. With the help of these improvements, ESRGAN got better visual quality and more realistic and natural textures.

\section{Conclusion}
\label{Conclusion}
Aiming at the topic of the application of deep learning in the field of image super-resolution, this paper introduces image super-resolution technology and traditional super-resolution reconstruction algorithms, and analyzes and summarizes the research direction of deep learning and its differences with traditional machine learning models. Through classification and introduction of image super-resolution algorithms based on deep learning, the development trend and research results of image super-resolution technology in recent years are presented. With the continuous improvement of deep neural network models, the application of deep learning in the field of image super-resolution will continue to emerge new designs and ideas, resulting in better image super-resolution algorithms with higher speed and higher image reconstruction quality.

\bibliography{main}
\bibliographystyle{plain}

\end{document}